\documentclass[letterpaper, 10 pt, conference]{ieeeconf}  

\IEEEoverridecommandlockouts

\usepackage{graphics} 
\usepackage{epsfig} 
\usepackage{mathptmx} 
\usepackage{times} 
\usepackage{amsmath} 
\usepackage{amssymb}  
\usepackage{subcaption}
\usepackage{multirow}
\usepackage[normalem]{ulem}
\useunder{\uline}{\ul}{}
\usepackage{pifont}
\newcommand{\cmark}{\ding{51}}
\newcommand{\xmark}{\ding{55}}
\usepackage{xcolor}

\usepackage{balance}
\usepackage{hyperref}
\usepackage[capitalise]{cleveref}

\title{\LARGE \bf
An Open-Source LiDAR and Monocular Off-Road Autonomous Navigation Stack
}

\author{Rémi Marsal$^{*1}$, Quentin Picard$^{*2}$, Adrien Poiré$^{1}$, Sébastien Kerbourc'h$^{1}$, Thibault Toralba$^{1}$, Clément Yver$^{1}$,\\Alexandre Chapoutot$^{1}$, David Filliat$^{2}$
\thanks{$^{*}$Equal contribution}
\thanks{$^{1}$U2IS, ENSTA, Institut Polytechnique de Paris, Palaiseau, France\newline
        {\tt\small remi.marsal@ensta-paris.fr}}
\thanks{$^{2}$AMIAD, Pôle Recherche, France\newline
        {\tt\small quentin.picard@polytechnique.edu}}
\thanks{All authors are members of LARIAD, a joint AMIAD-ENSTA laboratory.}
}

\begin{document}

\maketitle
\thispagestyle{empty}
\pagestyle{empty}

\begin{abstract}

Off-road autonomous navigation demands reliable 3D perception for robust obstacle detection in challenging unstructured terrain. While LiDAR is accurate, it is costly and power-intensive. Monocular depth estimation using foundation models offers a lightweight alternative, but its integration into outdoor navigation stacks remains underexplored.
We present an open-source off-road navigation stack supporting both LiDAR and monocular 3D perception without task-specific training. For the monocular setup, we combine zero-shot depth prediction (Depth Anything V2) with metric depth rescaling using sparse SLAM measurements (VINS-Mono). Two key enhancements improve robustness: edge-masking to reduce obstacle hallucination and temporal smoothing to mitigate the impact of SLAM instability.
The resulting point cloud is used to generate a robot-centric 2.5D elevation map for costmap-based planning. Evaluated in photorealistic simulations (Isaac Sim) and real-world unstructured environments, the monocular configuration matches high-resolution LiDAR performance in most scenarios, demonstrating that foundation-model-based monocular depth estimation is a viable LiDAR alternative for robust off-road navigation. By open-sourcing the navigation stack and the simulation environment, we provide a complete pipeline for off-road navigation as well as a reproducible benchmark. Code available at \url{https://github.com/LARIAD/Offroad-Nav}.

\end{abstract}

\begin{figure}[t]
  \centering
  \includegraphics[width=0.98\columnwidth]{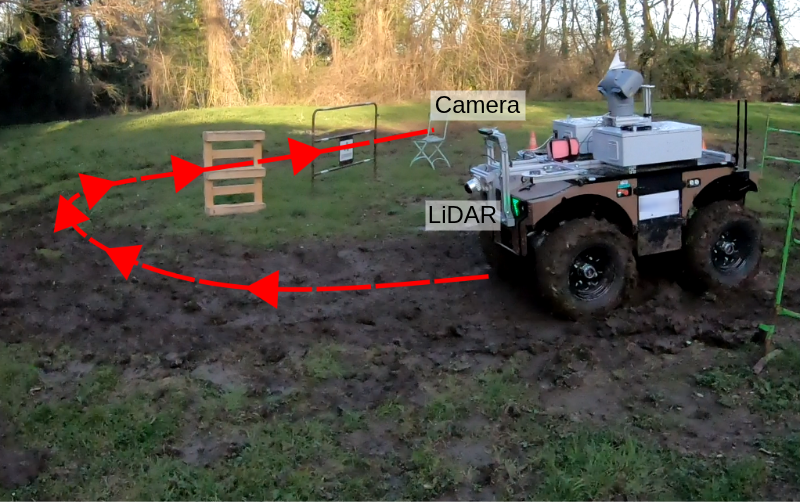}
  \vfill
  \vspace{0.2cm}
  \includegraphics[width=0.98\columnwidth]{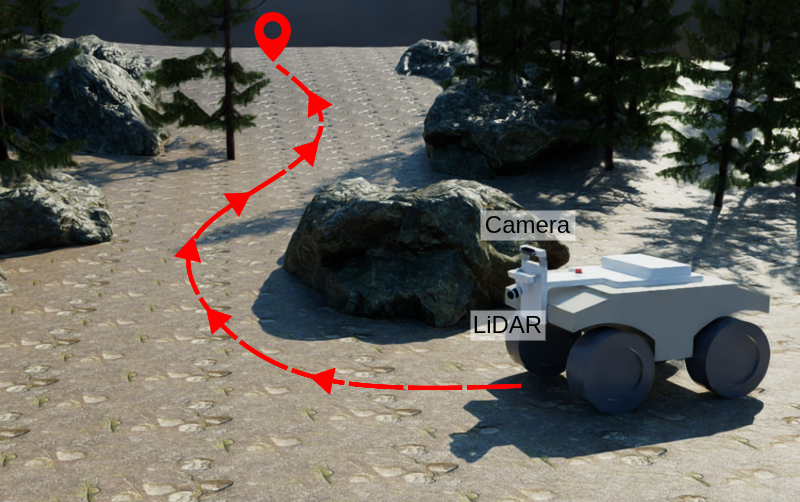}
  \caption{Autonomous navigation with the wheeled ground robot in the real and simulation off-road environment.}
  \label{fig:barakuda}
\end{figure}

\section{INTRODUCTION}

\begin{figure*}[ht]
  \centering
  \includegraphics[width=\textwidth]{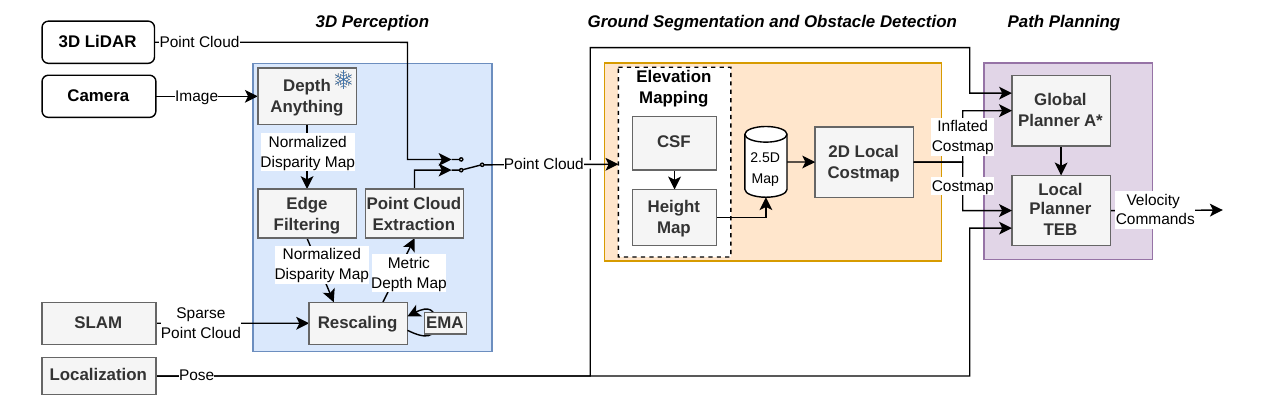}
  \caption{Diagram of our navigation pipeline. It takes as input either a LiDAR point could or monocular camera images with a SLAM sparse point cloud as well as the robot localization. It returns the velocity commands run by the robot.}
  \label{fig:stacknav}
\end{figure*}

Generating accurate 3D representations of the environment is crucial for off-road autonomous navigation.
Most approaches rely on LiDAR for this task due to its high accuracy.
However, LiDARs suffer from multiple disadvantages including their high cost, their ease of detection when discretion is required and their high energy consumption compared to cameras (typically 20W for LiDARs vs 2W for cameras).
Stereo depth cameras are frequently employed as an alternative to mitigate the aforementioned issues.
Yet, stereo vision systems face their own drawbacks; they notoriously struggle with low-textured regions, transparent objects or small details.
They are also sensitive to glare, low light or shadow and suffer from a limited depth range.

Numerous zero-shot learning-based monocular depth estimation models have been developed thanks to extensive training on huge datasets.
The different types of approaches to recover metric depth include zero-shot monocular metric depth estimation, zero-shot depth completion and depth rescaling.
Zero-shot monocular metric depth estimation~\cite{piccinelli2024unidepth, hu2024metric3d} consists in inferring a metric depth map directly from a single camera image.
Zero-shot depth completion methods~\cite{viola2024marigolddc, lin2025prompting} fuse an input image with sparse depth measurements to generate a metric depth map.
Depth rescaling techniques~\cite{marsal2025simple, guo2025monocular} estimate scaling and shift parameters to recover metric depth from normalized predictions produced by zero-shot monocular foundation models.
These approaches have drawbacks including hallucinations, poor generalization or temporal inconsistency if tested on very different scenes relative to their training set. They may also be very computationally expensive.
While these methods have found applications in indoor robotics \cite{guo2025igl, sun2025openfrontier}, their potential for off-road navigation remains largely unexplored.

Since all the previously discussed 3D perception approaches present both advantages and drawbacks, choosing one or another must depend on specific application contexts or constraints. 
This is why in this paper, we introduce an off-road navigation pipeline that can use either LiDAR or monocular depth estimation for 3D perception.
When using monocular depth estimation, we leverage a depth rescaling method \cite{marsal2025simple} in which the sparse metric depth is given by an inertial-visual SLAM to get a metric depth map before converting it into a point cloud.
To make the predicted depth maps usable in autonomous navigation, we propose a posthoc adaptation that improves discrimination between obstacles and navigable space, as well as a temporal smoothing of the scaling parameters.
Then, regardless of how the point cloud representing the scene is obtained, a cloth simulation filter (CSF)~\cite{zhang2016CSF} is applied in order to segment the ground from the positive obstacles.
This provides a height map that is converted into a costmap used for path planning and generating the robot commands.
Thus, by leveraging a foundation model for monocular depth prediction, our pipeline does not need additional learning or fine-tuning, which eases deployment in any environment.
We conducted extensive tests in complex off-road environments both simulated and real where the monocular perception demonstrated competitive results compared to the baseline with high-resolution LiDAR.
To facilitate further research and fair comparison, our navigation pipeline and the simulation environments will be made publicly available upon acceptance. 
To summarize, our contributions are the following:
\begin{enumerate}
    \item A ready-to-use open-source stack for off-road reactive navigation that requires no additional training to adapt to a new robotic platform or environment,
    \item The integration of monocular depth estimation for 3D perception in addition to LiDAR,
    \item Extensive experiments for evaluating our navigation stack on real and simulated environments, the latter of which are made publicly available upon acceptance.
\end{enumerate}

\section{RELATED WORK}

\subsection{3D perception for Off-Road Autonomous Navigation}

Most off-road autonomous navigation approaches perform 3D perception using a LiDAR \cite{elnoor2024amco, ruetz2024foresttrav, frey2024roadrunner} to generate an accurate point cloud of the surrounding environment.
Stereo vision is a relevant alternative to avoid LiDAR \cite{jung2024v, sivaprakasam2025salon} but it comes at a cost since the range and the robustness are reduced.
Several approaches propose to mitigate stereo depth drawbacks by learning a depth completion module \cite{meng2023terrainnet, zhang2025creste}.
In \cite{chung2024pixel}, the authors bypass the usual intermediate point cloud representation of the 3D geometry of the scene by directly reconstructing a $2.5$D elevation map from $3$ onboard monocular cameras.
However, these approaches are constrained by their reliance on learning frameworks tailored to specific robotic platforms and fixed sensor configurations. Consequently, even minor modifications necessitate extensive data recollection and retraining.
To address this limitation, our approach leverages foundation models and geometric methods as core components enabling robust generalization across diverse robotic systems and sensor configurations.

\subsection{Open-Source Off-Road Autonomous Navigation Pipeline}

Compared to autonomous driving \cite{jung2025open}, there exist few complete open-source stacks for autonomous off-road navigation.
The only complete off-road navigation stack is NATURE \cite{goodin2024nature} but it only supports LiDAR sensors for 3D perception and lacks rigorous evaluation of its navigation performance.
As most works focus on a small part of a whole navigation pipeline, only the code of the improved component is released, if ever made publicly available. 
It is worth noting that a more generic navigation stack such as the Nav2 stack \cite{macenski2020marathon2} can also be leveraged but it must be adapted for off-road navigation.
In this paper,  we present an off-road navigation stack that handles both LiDAR and monocular depth prediction for 3D perception.
We also conducted extensive experiments both in complex simulations and real environments to evaluate the ability to reach waypoints while avoiding obstacles.

\section{METHOD}

\Cref{fig:stacknav} details the proposed navigation pipeline. It is divided into three modules: the 3D perception module, the ground segmentation and obstacle detection module, and the path planning module.
The 3D perception module takes as input either a LiDAR point cloud or an image from a monocular camera and the sparse point clouds returned by a SLAM. 
While the LiDAR point cloud does not need to be processed by this module, the image and the sparse point cloud are leveraged to generate a much denser point cloud representation of the scene. 
The ground segmentation and obstacle detection module filters the ground from the input point cloud and estimates obstacles as an elevation map that is then converted into a binary costmap. Using this costmap and the robot localization, the path planning module estimates the optimal path to reach the goal with a global planner while the local planner provides the velocity commands. The localization is given by a classical data fusion through an extended Kalman filter based on GNSS, IMU data and SLAM pose estimates.

\subsection{3D Perception}

The 3D perception of the environment is either directly provided by a LiDAR or obtained through metric depth estimation from monocular images. 
For the latter use case, we leverage \cite{marsal2025simple} approach which consists of rescaling the relative depth maps given by a zero-shot monocular depth estimation model.
The sparse depth used for estimating the scaling and shift parameters is given by a visual-inertial SLAM approach (see \cref{sec:implementation} for details on the algorithm used).
\par
\begin{figure}[h]
  \centering
  \includegraphics{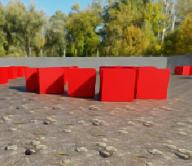}
  \vfill
  \vspace{0.2cm}
  \includegraphics{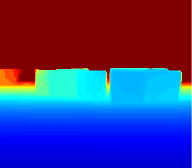}
  \hspace{0.cm}
  \includegraphics{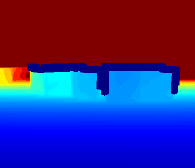}
  \vfill
  \vspace{0.2cm}
  \includegraphics{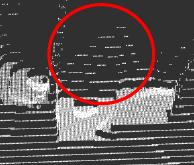}
  \hspace{0.cm}
  \includegraphics{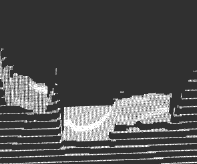}

  \caption{Estimated rescaled depth maps and the extracted $3$D point clouds from the top RGB image. The left side shows results without edge filtering, while the right side shows results with edge filtering applied.}
  \label{fig:monoedges}
\end{figure}

Due to blurry object borders in the depth map, the 3D back-projection of the depth map may result in an environment containing large phantom obstacles that may block the robot. 
This is illustrated in \cref{fig:monoedges} (left side) where the 3D points extracted from the estimated depth map contain outliers that can be seen in the red circle. The estimated depth map illustrated above shows the blurry object borders.
To address this issue, we filter the depth map posthoc by removing object edges. We apply a Sobel filter to the disparity map returned by the depth estimation model so as to only detect edges caused by depth discontinuities, especially those due to close objects. We then mask pixels within a five-pixel distance around the detected edges.
This removes phantom obstacles and makes it possible for the robot to pass behind obstacles. (see \cref{fig:monoedges}, right side).

\par
Instead of using the raw scaling and shift parameters $\hat{\mathbf{s}}_t$ estimated by the rescaling module at each timestep $t$, we smooth these parameters with the ones obtained at previous rescaling iteration using an exponential moving average:
\begin{equation}
    \tilde{\mathbf{s}}_t = \alpha\tilde{\mathbf{s}}_{t-1} + (1 - \alpha) \hat{\mathbf{s}}_t.
\end{equation}
The vector $\tilde{\mathbf{s}}_t$ stores the final scale and shift parameters that will be used to generate the metric depth map.
This improves robustness against the SLAM instabilities.

\subsection{Navigation pipeline}

The $3$D point cloud is used to represent the surrounding obstacles in an elevation map \cite{miki2022elevation}. In order to have a robust $2.5$D map in the context of off-road navigation where terrain may not be flat, the \emph{cloth simulation filter} method (CSF) \cite{zhang2016CSF} is used to separate ground points from non-ground points. The CSF method inverts the point cloud and simulates a rigid cloth falling onto the inverted surface to approximate the ground. Note that the cloth’s rigidity is an adjustable parameter that can be tuned based on the environment, see \cref{sec:configuration-navigation-pipeline}. The height of the obstacles, corresponding to the distance between the points and the cloth surface, is stored in a robot-centric $2.5$D grid map, the elevation map.
A memory buffer is implemented to store the positions of each new and already known obstacle within the size limit of the grid map. These obstacles are shifted according to the movement of the robot. This allows the path planning to be more robust, as obstacles that leave the sensor's field of view are not immediately forgotten but persist in the map until they are updated by new observations.
\par
The path planning is based on a \emph{costmap} generated from the aforementioned grid map using a height threshold. Multiple thresholds can be used to classify the obstacles at different levels. In the results of this paper, a 30cm threshold is used.
An $A^*$ global planner is used on the costmap that has been inflated according to the robot size. It provides the high-level path to follow for the \emph{Timed-Elastic-Bands} (TEB) local planner \cite{Rsmann2027teb} which computes a flexible path planning for better robustness to perception errors. 

\section{EXPERIMENTS}

Our navigation stack is evaluated in both simulated and real-world environments, across a range of difficulty levels. In each scenario, the robot’s mission is to autonomously reach a specified goal position while dynamically avoiding obstacles. Importantly, the robot does not rely on pre-programmed waypoints to complete its task.

\subsection{Evaluation benchmark}

\subsubsection{Simulated environments} We created three different simulation environments (see \cref{fig:quali_results_simu} for top view) using Isaac Sim \cite{isaacsim}. Each environment corresponds to a difficulty level: easy, medium or hard. 
The easy environment consists of a flat terrain with red cubes that have been placed by hand either alone or in clusters to create bigger obstacles.
The medium environment leverages photorealistic obstacles, trees and rocks, instead of the red cubes.
The hard environment includes the same assets as the medium one but it also adds ground elevation and high grass as traversable obstacles.
For each simulated environment, we define three scenarios with the same starting position (at coordinate $(0,0)$ in \cref{fig:quali_results_simu}) and three different goals located at $10$, $20$, and $30$ meters as the crow flies and indicated (with an orange, a green and a blue star, respectively, in \cref{fig:quali_results_simu}).

\subsubsection{Real scenarios} We create three different scenarios with increasing difficulty level for evaluating the navigation capacities of our stack in real off-road environments (see \cref{fig:quali_results_real}, obstacles are in white).
The easy scenario consists of a single obstacle positioned in the middle of the straight line between the starting point and the goal on a flat terrain.
The medium scenario starts and ends in the same location as the easy scenario, but it has more obstacles which force the robots to take more detours.
Finally, the hard scenario is the longest: in addition to obstacles on a flat terrain, it begins on a road at the bottom of a slope with obstacles.

\subsection{Metrics}\label{sec:metrics}

We evaluate the performance of our navigation pipeline using three metrics: the \emph{Success Rate}, the \emph{Success weighted by Path Length} and the \emph{Distance Ratio}.
The Success Rate (SR) measures the proportion of runs in which the robot reached the goal (see \cref{eq:sr_metric}). The Success weighted by Path Length (SPL) quantifies the path efficiency relative to a reference trajectory when the run is a success (see \cref{eq:spl_metric}). The Distance Ratio (DR) gives the maximum progress of the robot towards the goal (see \cref{eq:dr_metric}). These metrics are computed for each scenario using the following formulas:
\begin{align}
     \text{SR} & = \frac{1}{N}\sum_{i=1}^{N} S_i 
     \label{eq:sr_metric}
     \\
     \text{SPL} & =\frac{1}{N}\sum_{i=1}^{N} S_i \frac{\ell^*}{\hat{\ell}_i}
     \label{eq:spl_metric}
     \\
     \text{DR}& = \frac{1}{N}\sum_{i=1}^{N}
        \begin{cases}
            1 & \text{if } S_i = 1 \\
            \displaystyle \max\limits_{d_{ij} \leq \delta} \ell^*_j / \ell^* & \text{otherwise.}
        \end{cases}
        \label{eq:dr_metric}
\end{align}

where $N$ is the total number of runs; $S_i$ and $\hat{\ell}_i$ are the binary success value and the length of the robot path from the start to the goal for the run $i$, respectively. We consider a successful run ($S_i = 1$) if the robot's final position is within $2$m of the goal.
We refer to $\ell^*$ as the total length of a reference trajectory for a given scenario, and $\ell^*_j$, for its length at the $j$-th step of the trajectory.
In practice, we have remotely operated the robot to generate the trajectories $\ell^*$ as efficiently as possible towards each goal in each environment (see dashed lines in \cref{fig:quali_results_simu} and \cref{fig:quali_results_real}).
We define $d_{ij}$ as the minimal distance between the robot location at run $i$ and timestep $j$ and any position of the reference trajectory.

\par
In simulation, $N$ is set to $10$ runs per goal distance for each environment, which corresponds to $90$ runs for each $3$D perception method. In the real-world environment, $3$ runs have been computed per scenario, which correspond to $9$ runs for each input modality.

\subsection{Implementation Details}\label{sec:implementation}

\subsubsection{Hardware details}
Our experiments in real environments are performed using a Barakuda robot from Shark Robotics that we equipped with an Ouster Dome LiDAR with $128$ channels, a ZED2i camera, an Ellipse D SBG IMU double GNSS antennas, a Jetson AGX Orin $64$GB as embedded computer and a WiFi router for distant communication (see \cref{fig:barakuda} top).
In simulation, we recreated a Barakuda platform only equipped with the same LiDAR and the same camera as the real one, the other sensor being unnecessary since the localization used is the one provided by the simulator (see \cref{fig:barakuda} bottom).

\begin{figure*}[htb]
    \centering
    
    \begin{subfigure}[b]{0.32\textwidth}
        \includegraphics[width=\textwidth]{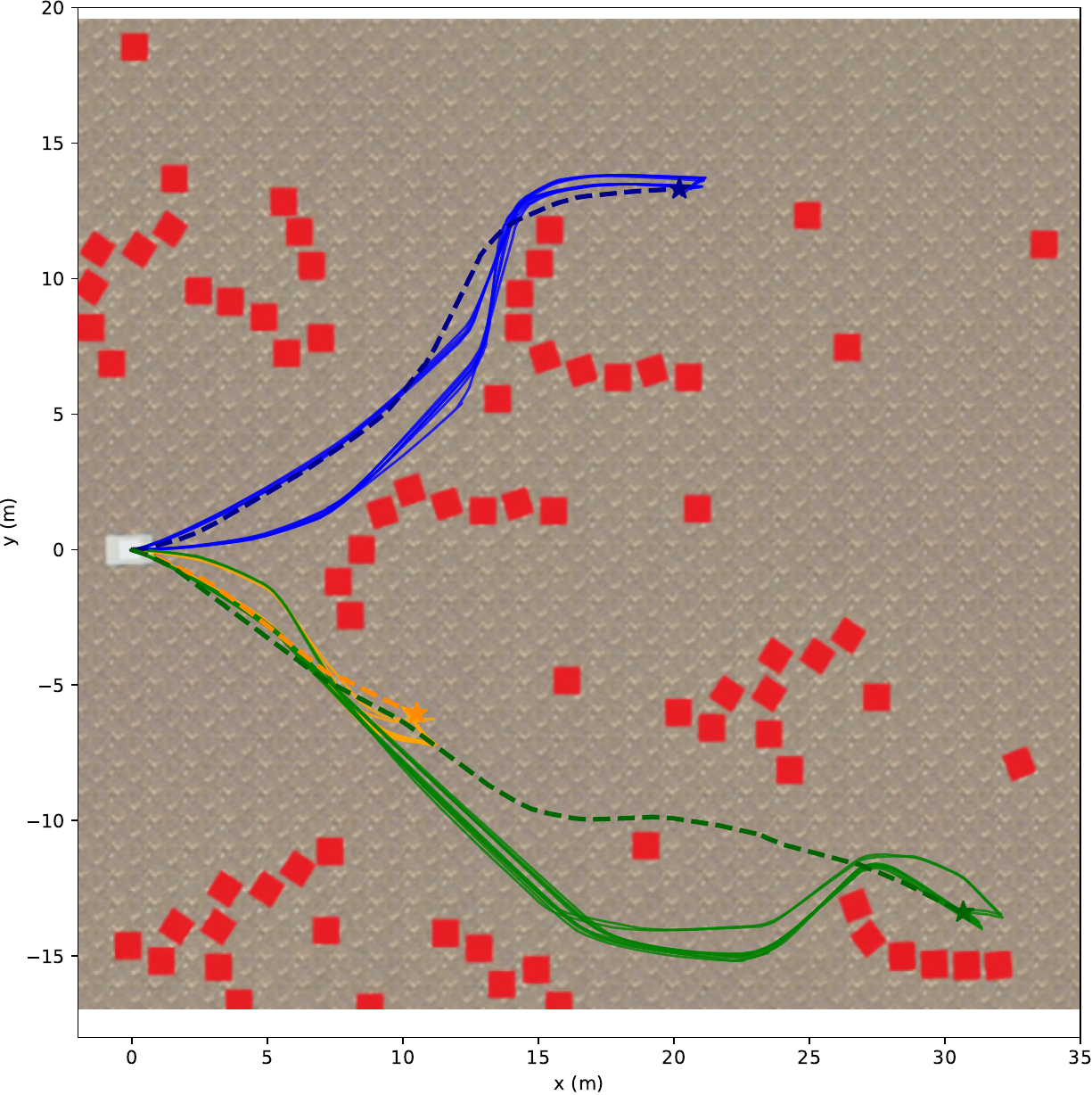}
    \end{subfigure}
    \hfill
    \begin{subfigure}[b]{0.32\textwidth}
\includegraphics[width=\textwidth]{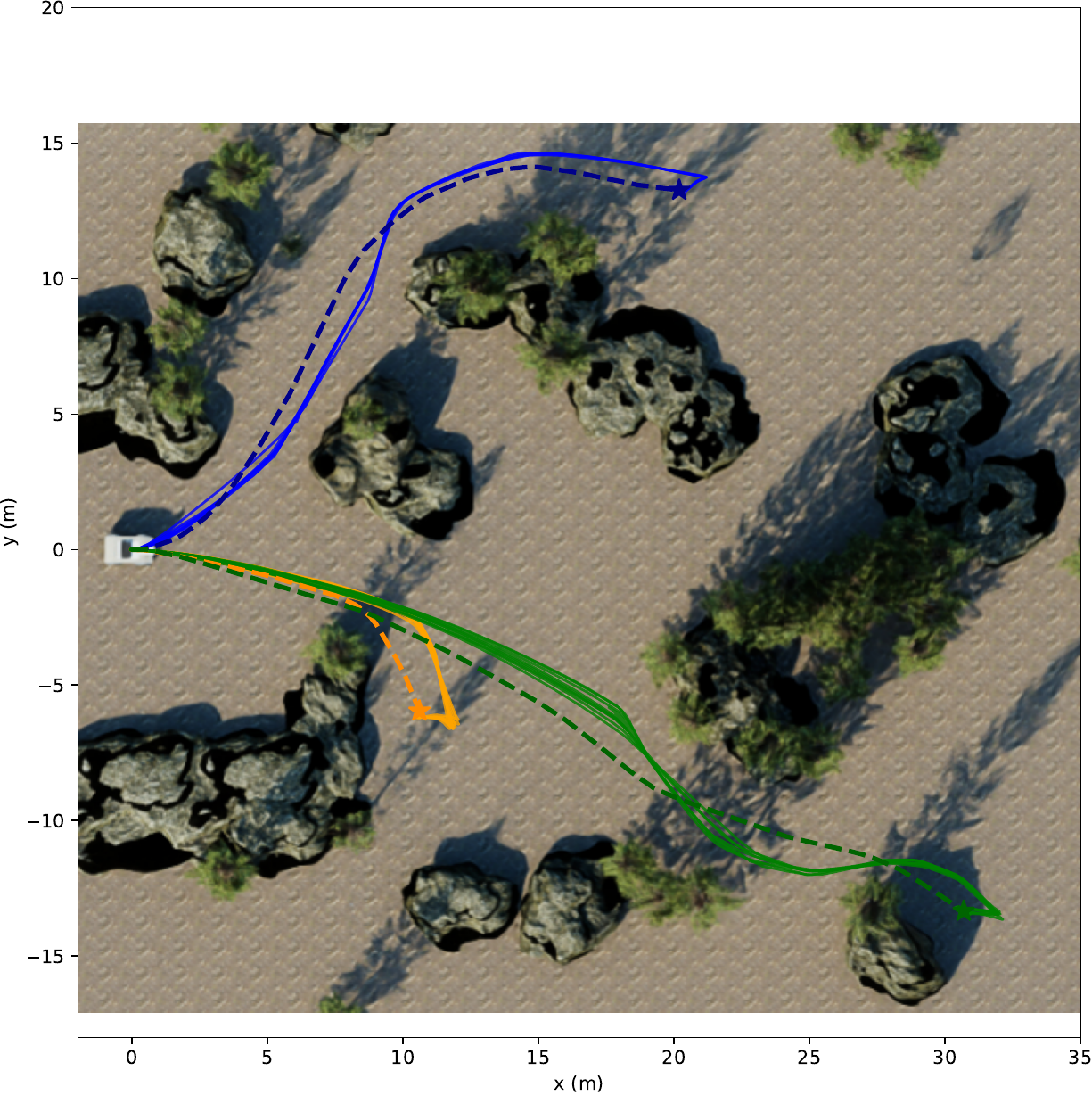}
    \end{subfigure}
    \hfill
    \begin{subfigure}[b]{0.32\textwidth}
        \includegraphics[width=\textwidth]{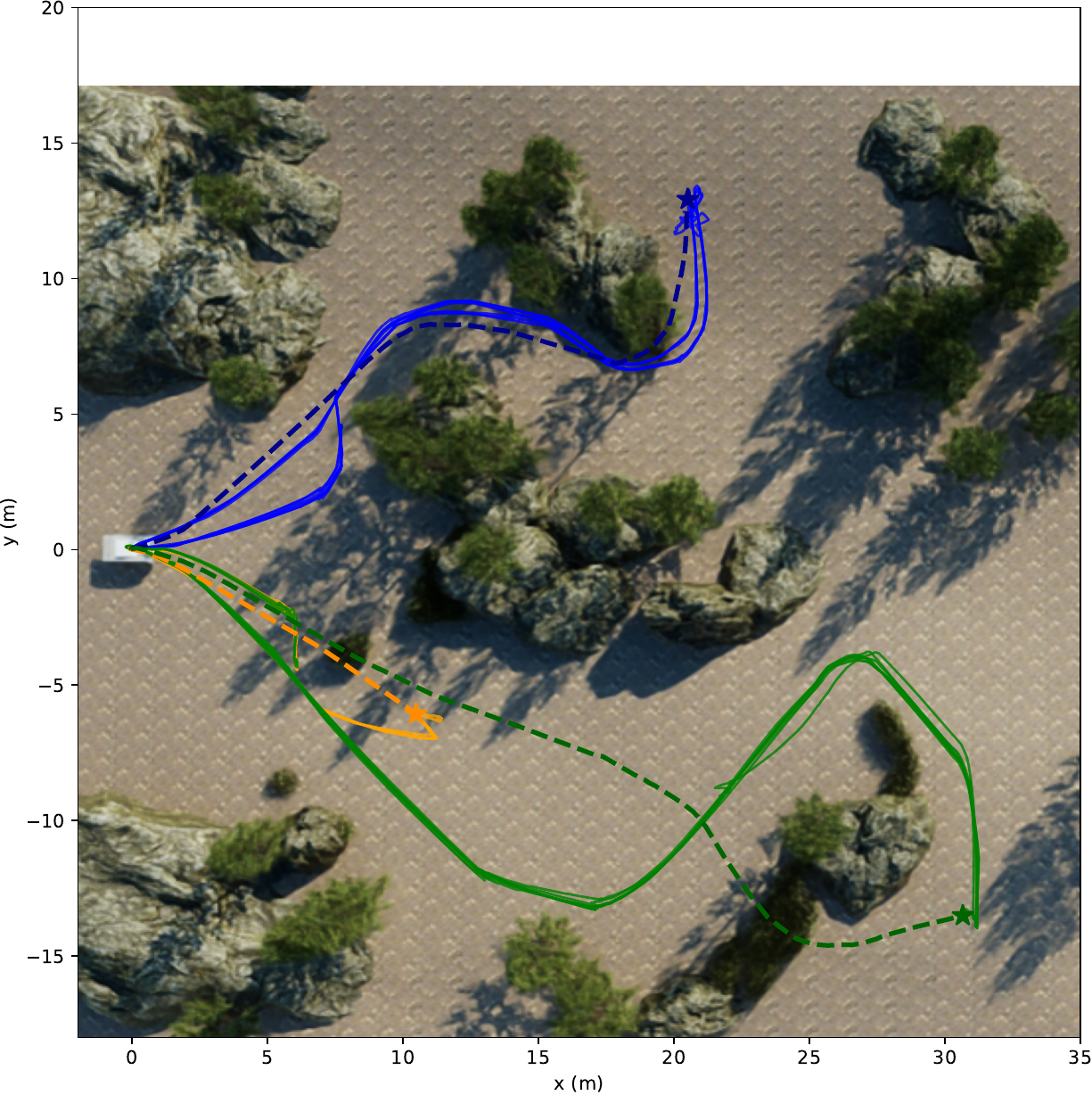}
    \end{subfigure}

    \vspace{0.5em}

    \begin{subfigure}[b]{0.32\textwidth}
        \includegraphics[width=\textwidth]{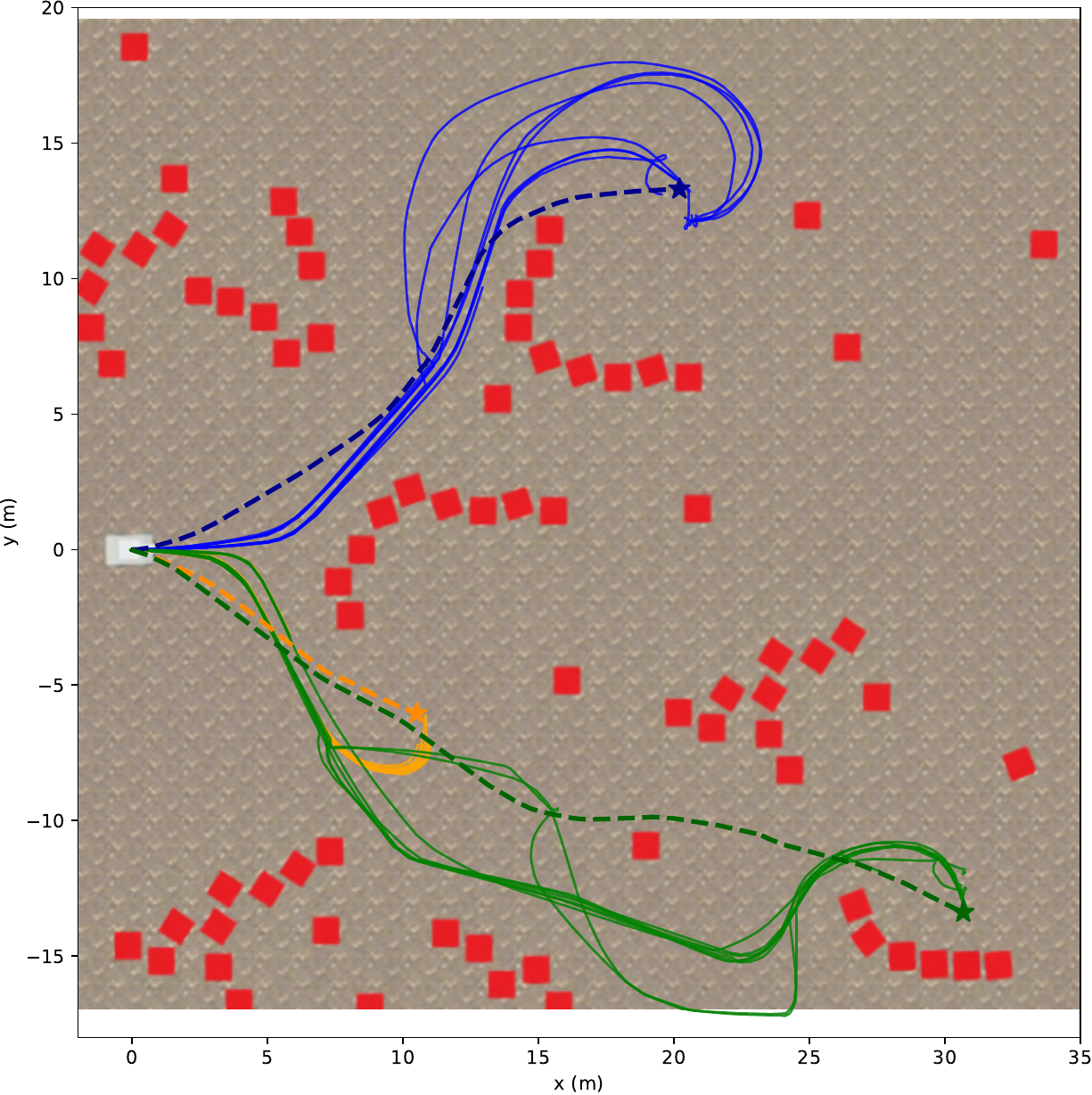}
    \end{subfigure}
    \hfill
    \begin{subfigure}[b]{0.32\textwidth}
        \includegraphics[width=\textwidth]{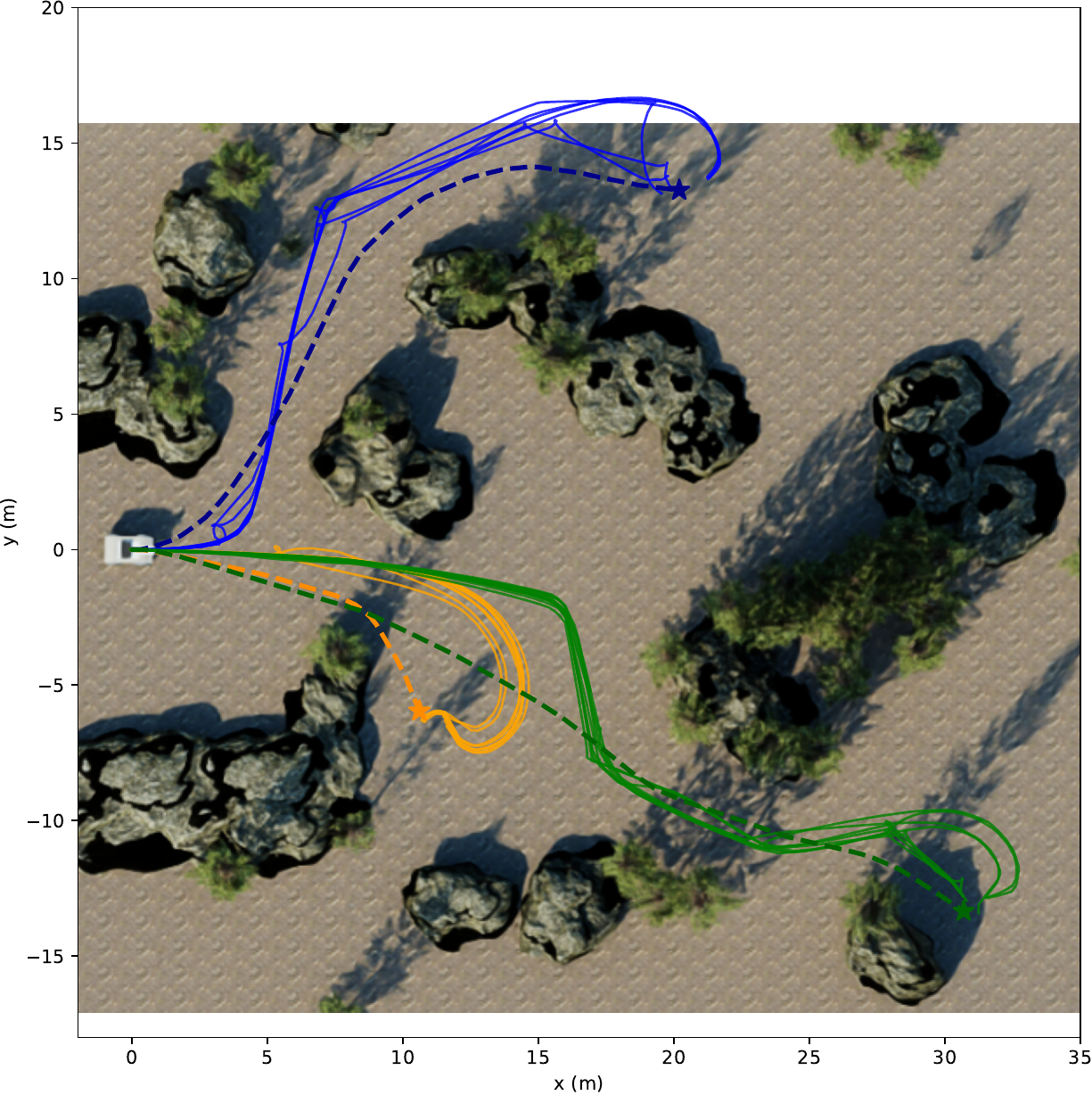}
    \end{subfigure}
    \hfill
    \begin{subfigure}[b]{0.32\textwidth}
        \includegraphics[width=\textwidth]{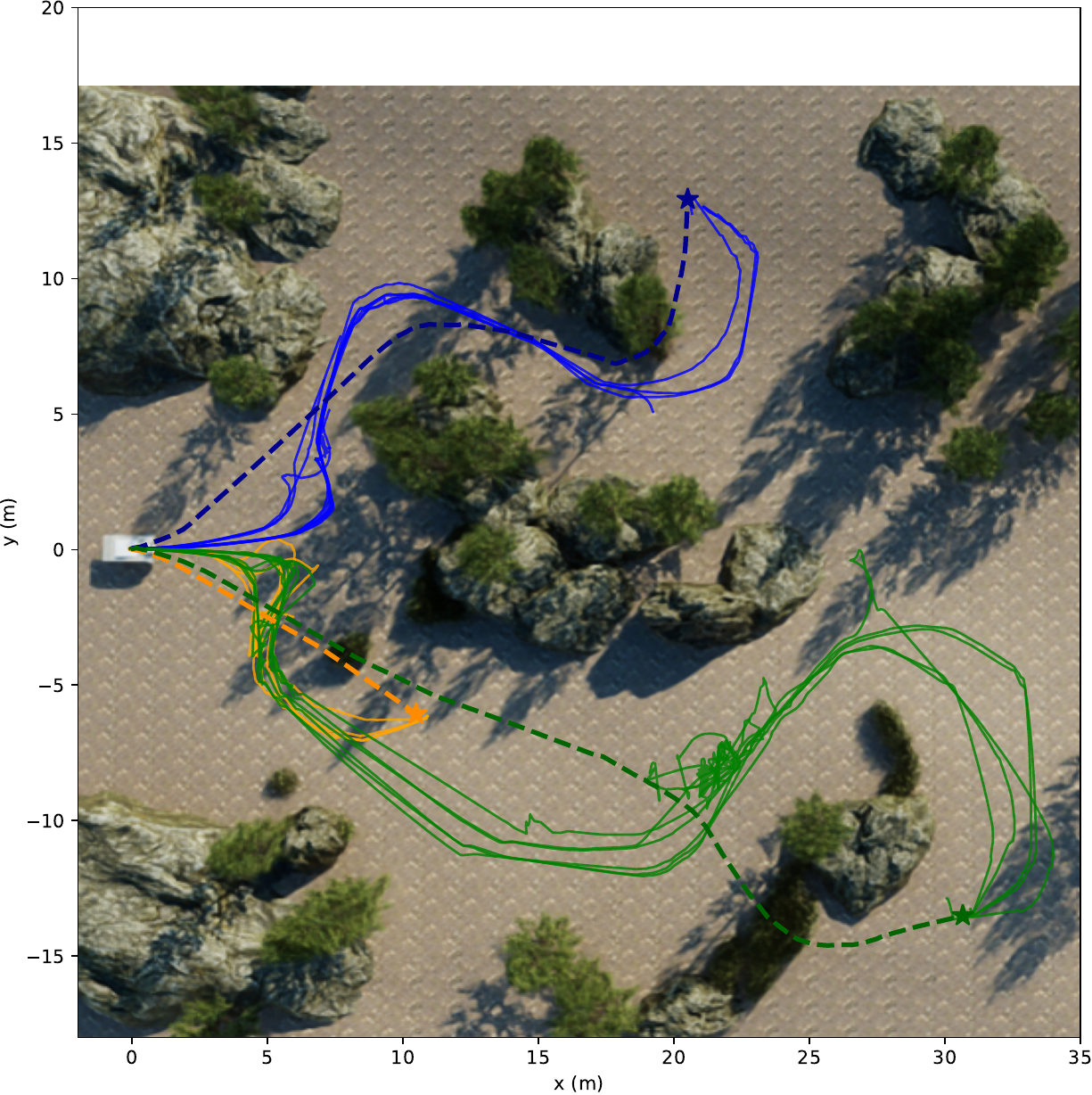}
    \end{subfigure}

    \vspace{0.5em}

    \begin{subfigure}[b]{0.32\textwidth}
        \includegraphics[width=\textwidth]{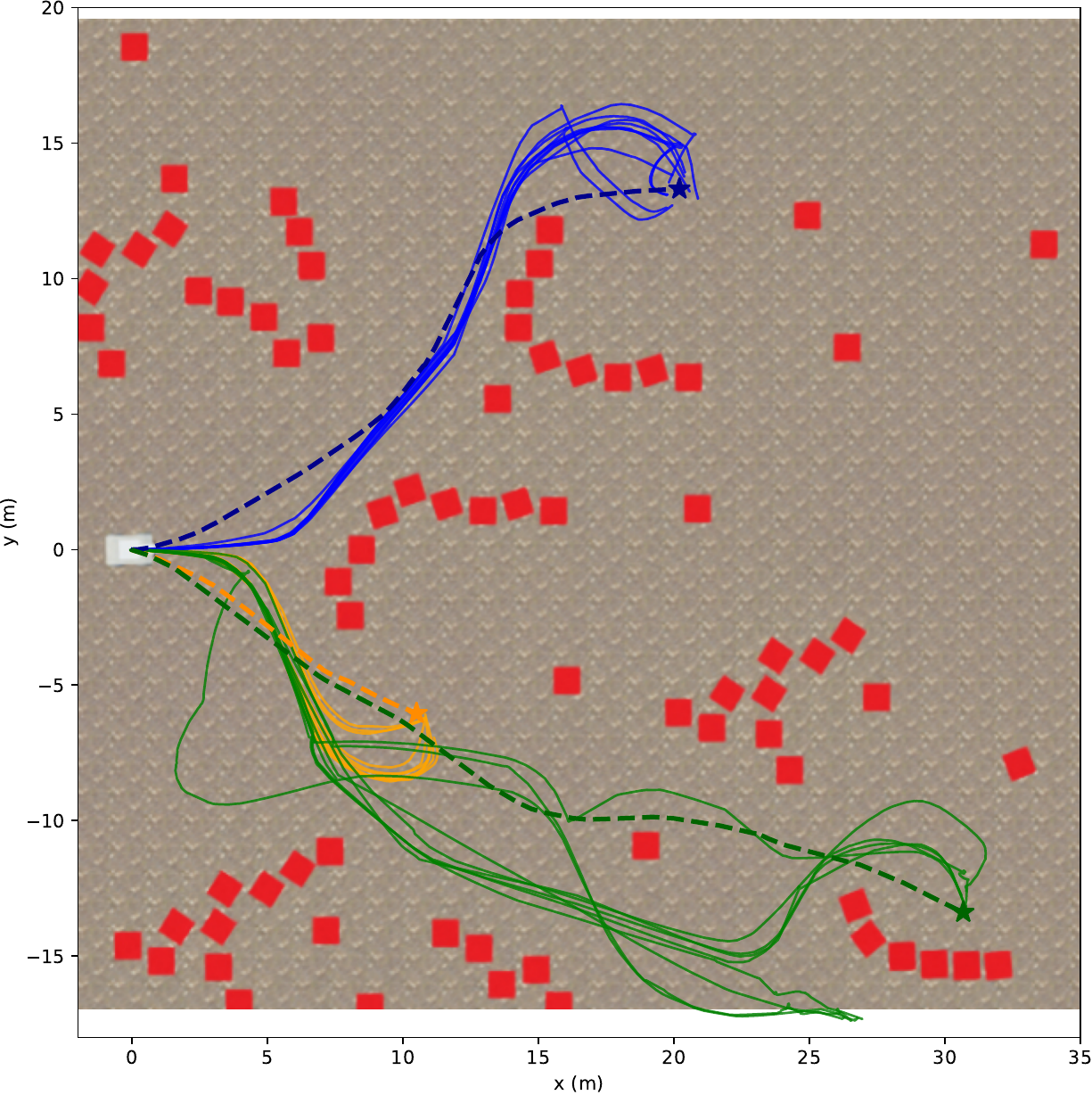}
        
    \end{subfigure}
    \hfill
    \begin{subfigure}[b]{0.32\textwidth}
        \includegraphics[width=\textwidth]{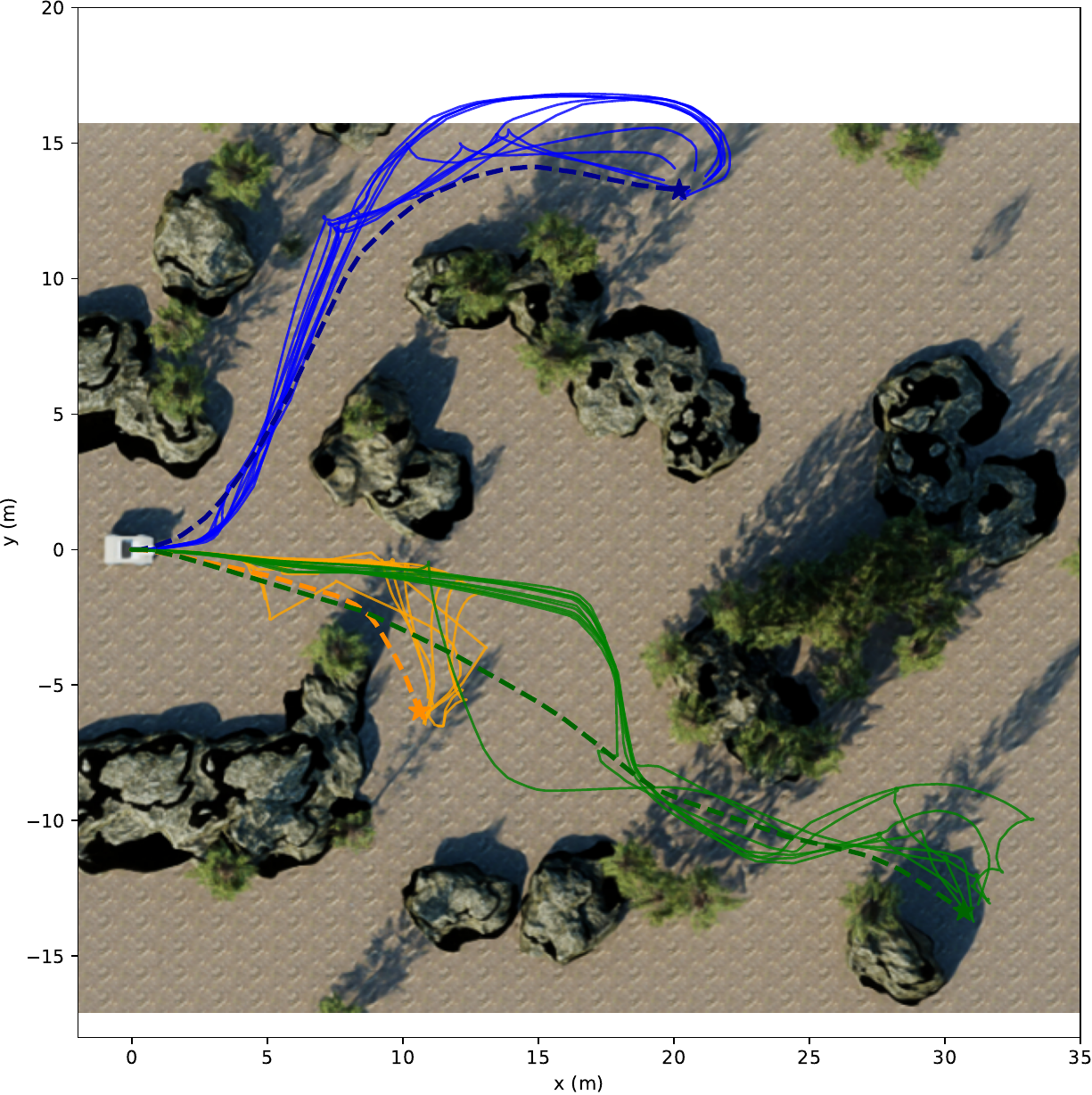}
        
    \end{subfigure}
    \hfill
    \begin{subfigure}[b]{0.32\textwidth}
        \includegraphics[width=\textwidth]{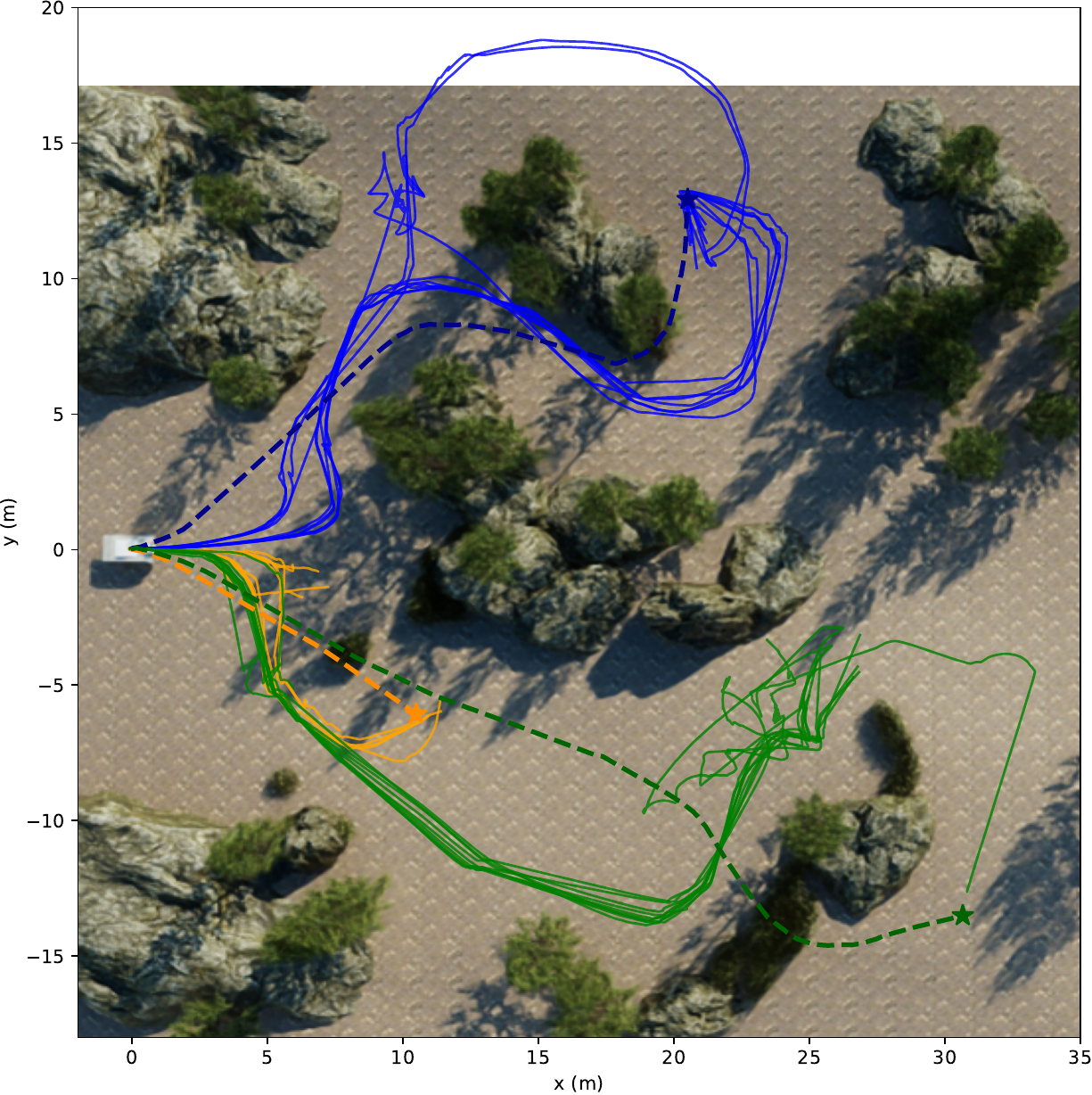}
        
    \end{subfigure}

    \caption{Top views of the simulation experiments. From left to right, the easy, medium and hard simulated environments, respectively. In each environment, the goal points of the $10$, $20$ and $30$ meters scenarios is indicated with an orange, blue and green star, respectively. The reference remotely operated trajectories are given in dashed lines and the ones performed by the robot are shown in solid lines. From top to bottom: the robot trajectories are obtained using the LiDAR (sim-tuned), the LiDAR (real-params) and Mono-VINS (STCD only), respectively.}
    \label{fig:quali_results_simu}
\end{figure*}

\subsubsection{Software implementation} 
Our navigation stack has been implemented with the ROS Noetic framework.

Regarding the monocular depth estimation, the normalized depth map is estimated using Depth Anything V2 \cite{yang2024DAV2} accelerated through FP16 quantization and TensorRT optimization. The reference point cloud used for rescaling is given by VINS-Mono \cite{qin2018vins} and the temporal smoothing parameter is set to $\alpha=0.8$.
It is worth noting that tracking image features accurately, as performed in VINS-Mono, is impossible in simulation due to the repetition of the same texture on the ground (see \cref{fig:quali_results_simu}) and temporal ghosting in the rendering of the vegetation.
Consequently, VINS-Mono cannot be used in simulation to provide the reference point cloud needed for rescaling the normalized depth map.
Instead, we sample points in the camera image with the \emph{Shi-Tomasi Corner Detector} (STCD) \cite{shi1994good} used by VINS-Mono and leverage the ground-truth depth at these locations as reference in the rescaling module. In the results, it is referred to as ``VINS-Mono (STCD only)".
In real scenarios, this issue does not arise, so VINS-Mono is normally used.
Regarding CSF, it contains parameters, including cloth properties and the optimization  process settings for the cloth simulation, that must be adjusted according to the environment.
The TEB local planner's parameters mainly depend on the robot characteristics and on optimization  process.
Due to space constraints, we do not detail CSF and TEB's parameters here.

\subsubsection{Configuration for the navigation pipeline}
\label{sec:configuration-navigation-pipeline}

Two configurations of the navigation stack are used in the experiments: \textit{sim-tuned} and \textit{real-params}.
The \textit{sim-tuned} settings is the ideal configuration in which the CSF parameters are adapted with a different rigidness for each the environment so the cloth deformation always converges optimally, and the local TEB planner is set to follow the global planner without any possible deviation.
However, this configuration is naive and cannot be used in real environment as it does not take into account potential perception or localization errors. For this reason, it is only used in simulation with the LiDAR to create a baseline.
The \textit{real-params} setting corresponds to the one used on the real robot. It is also applied in simulation with the LiDAR and the monocular depth estimation.
In this configuration, the CSF parameters are the same regardless the environment.
The trajectory optimized by the local planner is constrained by the global planner and a penalization when the robot gets too close to obstacles, making the navigation much more resilient to perception or localization errors.

\subsection{Qualitative results}\label{sec:quali}

\begin{figure*}[htbp]
    \centering
    
    \begin{subfigure}[b]{0.32\textwidth}
        \includegraphics[width=\textwidth]{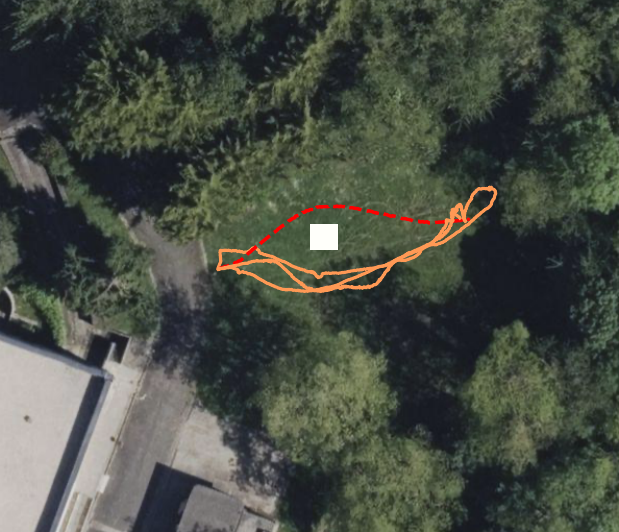}
    \end{subfigure}
    \hfill
    \begin{subfigure}[b]{0.32\textwidth}
        \includegraphics[width=\textwidth]{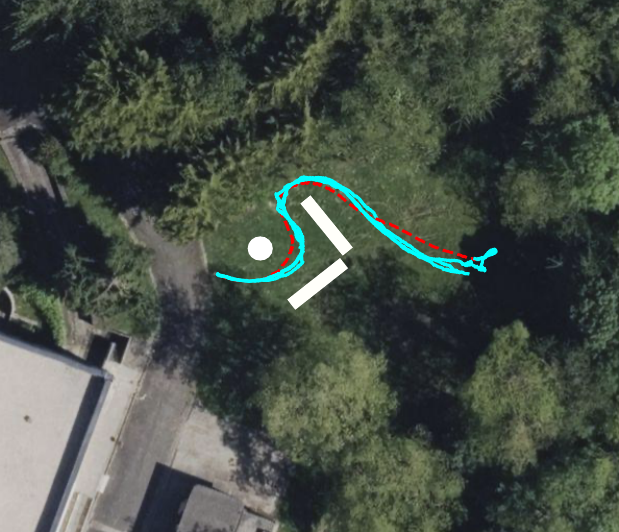}
    \end{subfigure}
    \hfill
    \begin{subfigure}[b]{0.32\textwidth}
        \includegraphics[width=\textwidth]{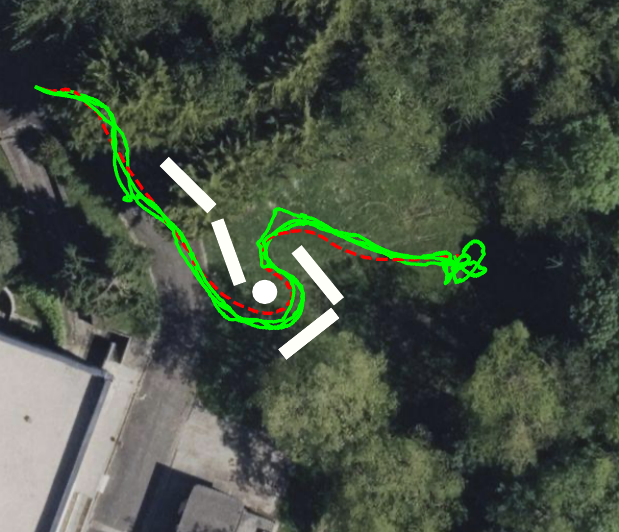}
    \end{subfigure}

    \vspace{0.5em}

    \begin{subfigure}[b]{0.32\textwidth}
        \includegraphics[width=\textwidth]{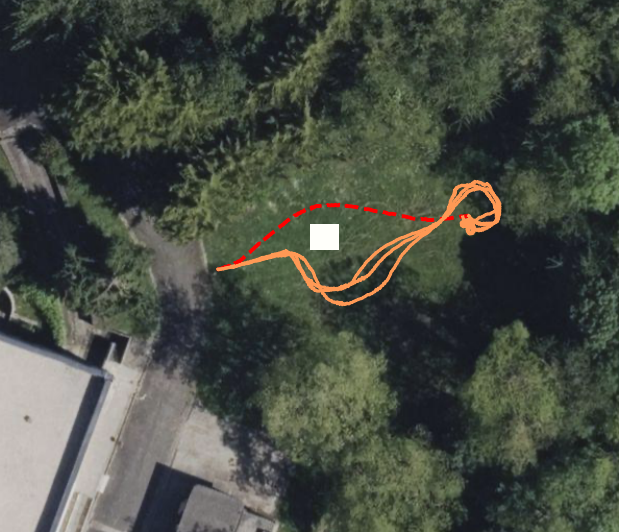}
        
    \end{subfigure}
    \hfill
    \begin{subfigure}[b]{0.32\textwidth}
        \includegraphics[width=\textwidth]{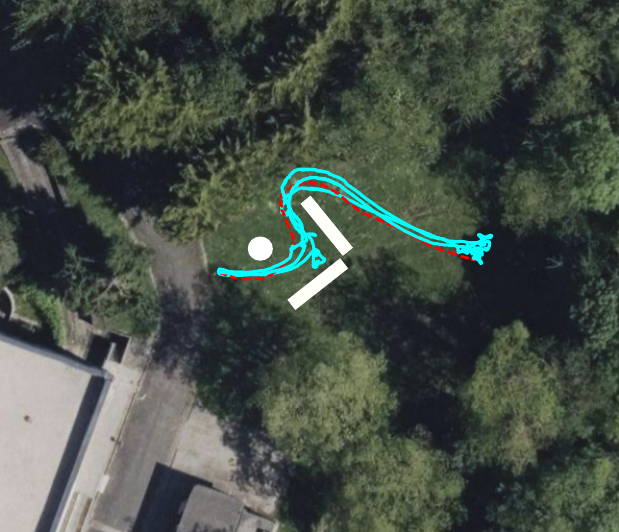}
        
    \end{subfigure}
    \hfill
    \begin{subfigure}[b]{0.32\textwidth}
        \includegraphics[width=\textwidth]{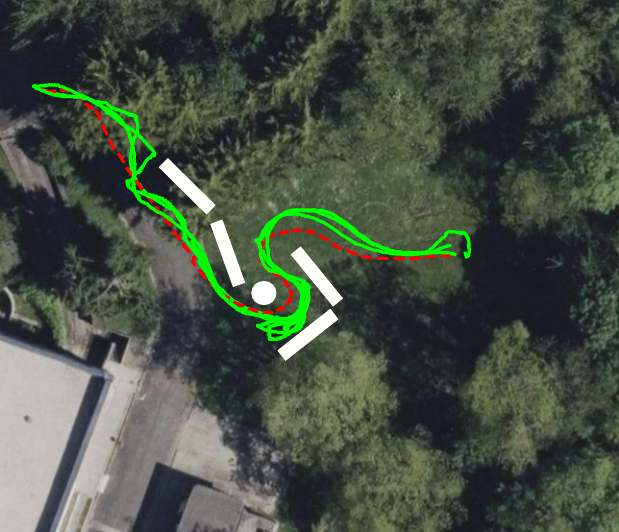}
        
    \end{subfigure}

    \caption{Top view for the real-world experimental area where obstacles are placed on an off-road terrain. The first row shows the LiDAR experiments and the second one, the runs with the monocular camera. From left to right: the easy, medium and hard scenario. Obstacles are represented in white circles and rectangles. The dashed lines correspond to the remotely operated reference trajectories.}
    \label{fig:quali_results_real}
\end{figure*}

\begin{table*}[ht]
\centering
\caption{Navigation performance across simulated scenarios and goal distances.}
\label{tab:simu_results}

\begin{tabular}{l|ccc|ccc|ccc|ccc}
\hline
\multirow{2}{*}{\textbf{}} & \multicolumn{3}{c|}{\textbf{Easy}} & \multicolumn{3}{c|}{\textbf{Medium}} & \multicolumn{3}{c|}{\textbf{Hard}} & \multicolumn{3}{c}{\textbf{Average}} \\
 & SR$\uparrow$ & SPL$\uparrow$ & DR$\uparrow$ & SR$\uparrow$ & SPL$\uparrow$ & DR$\uparrow$ & SR$\uparrow$ & SPL$\uparrow$ & DR$\uparrow$ & SR$\uparrow$ & SPL$\uparrow$ & DR$\uparrow$ \\
\hline
\multicolumn{13}{c}{\textit{10m goal}} \\
\hline
LiDAR (sim-tuned)  & \textbf{1.00} & 0.70 & \textbf{1.00} & \textbf{1.00} & \textbf{0.73} & \textbf{1.00} & 0.40 & 0.31 & 0.76 & \textbf{0.80} & \textbf{0.58} & \textbf{0.92} \\
LiDAR (real-params)  & 0.90 & 0.66 & 0.95 & 0.80 & 0.47 & 0.93 & 0.40 & 0.29 & 0.72 & 0.70 & 0.47 & 0.87 \\

Mono-VINS (STCD only) & \textbf{1.00} & \textbf{0.73} & \textbf{1.00} & 0.80 & 0.59 & 0.89 & \textbf{0.50} & \textbf{0.39} & \textbf{0.78} & 0.77 & 0.57 & 0.89 \\
\hline
\multicolumn{13}{c}{\textit{20m goal}} \\
\hline
LiDAR (sim-tuned)  & 0.90 & 0.78 & 0.96 & 0.40 & 0.35 & 0.56 & 0.60 & 0.45 & 0.76 & 0.63 & 0.53 & 0.76 \\
LiDAR (real-params)  & 0.80 & 0.60 & 0.90 & 0.60 & 0.49 & 0.69 & 0.30 & 0.22 & 0.62 & 0.57 & 0.44 & 0.74 \\

Mono-VINS (STCD only)  & \textbf{1.00} & \textbf{0.83} & \textbf{1.00} & \textbf{1.00} & \textbf{0.83} & \textbf{1.00} & \textbf{0.90} & \textbf{0.58} & \textbf{0.93} & \textbf{0.97} & \textbf{0.75} & \textbf{0.98} \\
\hline
\multicolumn{13}{c}{\textit{30m goal}} \\
\hline
LiDAR (sim-tuned)  & \textbf{0.80} & \textbf{0.66} & \textbf{0.94} & 0.70 & 0.58 & 0.82 & \textbf{0.70} & \textbf{0.47} & \textbf{0.76} & \textbf{0.73} & \textbf{0.57} & \textbf{0.84} \\
LiDAR (real-params)  & 0.70 & 0.55 & 0.83 & \textbf{0.80} & \textbf{0.68} & \textbf{0.91} & 0.50 & 0.30 & 0.65 & 0.67 & 0.51 & 0.80 \\

Mono-VINS (STCD only)  & 0.50 & 0.41 & 0.70 & 0.70 & 0.59 & 0.79 & 0.10 & 0.05 & 0.70 & 0.43 & 0.35 & 0.73 \\
\hline
\end{tabular}

\end{table*}

\begin{table*}[ht]
\centering
\caption{Navigation performance across real scenarios with the Barakuda robot.}
\label{tab:real_results}

\begin{tabular}{l|ccc|ccc|ccc|ccc}
\hline
\multirow{2}{*}{\textbf{}} & \multicolumn{3}{c|}{\textbf{Easy}} & \multicolumn{3}{c|}{\textbf{Medium}} & \multicolumn{3}{c|}{\textbf{Hard}} & \multicolumn{3}{c}{\textbf{Average}} \\
 & SR$\uparrow$ & SPL$\uparrow$ & DR$\uparrow$ & SR$\uparrow$ & SPL$\uparrow$ & DR$\uparrow$ & SR$\uparrow$ & SPL$\uparrow$ & DR$\uparrow$ & SR$\uparrow$ & SPL$\uparrow$ & DR$\uparrow$ \\
\hline
LiDAR (real-params)  & \textbf{1.00} & \textbf{0.73} & \textbf{1.00} & \textbf{1.00} & \textbf{0.68} & \textbf{1.00} & \textbf{1.00} & \textbf{0.67} & \textbf{1.00} & \textbf{1.00} & \textbf{0.69} & \textbf{1.00} \\
Mono-VINS & \textbf{1.00} & 0.51 & \textbf{1.00} & \textbf{1.00} & 0.63 & \textbf{1.00} & \textbf{1.00} & 0.63 & \textbf{1.00} & \textbf{1.00} & 0.59 & \textbf{1.00} \\
\hline
\end{tabular}

\end{table*}

The qualitative results for the simulated environments and the real scenarios are shown in \cref{fig:quali_results_simu} and \cref{fig:quali_results_real}, respectively.

Each row of the simulation results corresponds to a different navigation pipeline configuration: the LiDAR (sim-tuned), the LiDAR (real-params) and then the VINS-Mono (STCD only). The trajectories of the LiDAR sim-tuned configuration show a low variability between runs, regardless of the environment, as the local planner is set to closely follow the global planner. The LiDAR real-params trajectories are much more diverse, especially in the easy environment with the $20$ meters goal or the hard one with the $30$ meters goal since the local planner has more flexibility relative to the global planner. The monocular configuration in the third row challenges the use of the LiDAR sensor for navigation with similar results for the easy and medium scenarios but more diverse trajectories are shown in the hard environment in blue. The latter, with the goal at $30$ meters in green, also shows that only one trajectory manages to reach the goal. Many runs are aborted as the TEB local planner did not manage to find a solution to generate a trajectory. The hardest environment in the third column also displays the behaviour of the navigation stack in the presence of high grass for each configuration in the $10$m and $30$m goal scenarios. Since it is detected as an obstacle, the estimated trajectory cannot pass through it as the reference trajectory does. The high grass is much more problematic for the monocular depth estimation than the trees and rocks, leading to many aborted runs. 
The monocular version struggles to find a feasible trajectory to avoid the high grass obstacles because they cause a larger blurry region in the depth map since they do not have sharp edges like rocks or cubes.
Therefore, the edge filtering struggles to detect and remove high grass obstacles' edges, leading to outliers $3$D points behind the real obstacles when the depth map is converted into a point cloud, which are interpreted as obstacles blocking the way to the goal.
On the contrary, the $20$m goal scenario does not contain any grass to avoid and most of the generated trajectories reach the goal. 

\par
The real-world experiments are shown in \cref{fig:quali_results_real} with the LiDAR at the top and the monocular camera at the bottom. Columns one to three correspond to the easy, medium and hard scenarios, respectively. The obstacles (see those shown in \cref{fig:barakuda}, top) are identified by the white circles and rectangles.
Overall, the trajectories generated using the monocular camera are similar to the ones obtained with the LiDAR. The monocular stack appears to turn later than with the LiDAR for avoiding obstacles, resulting in less efficient trajectories. This comes from a longer information processing chain due to the monocular depth estimation (see \cref{sec:exec}). Furthermore, the trajectories of the robot exhibit back-and-forth movements to avoid the circular obstacle when using the monocular configuration in the medium and hard scenarios. This behavior, caused by path planning instabilities, can be mainly attributed to the limited field of view and depth estimation accuracy of the monocular camera, in addition to the aforementioned embedded computational issues. In contrast, the LiDAR-based configuration produces smoother trajectories because of less computational load on the processor (see \cref{sec:exec}). It allows the planner to anticipate obstacles earlier and generate more efficient avoidance maneuvers.
The observed difficulty of our robot in reaching the final position when it gets closer to it stems from the TEB local planner configuration, indicating that further parameter optimization is necessary.

\subsection{Quantitative results}

The quantitative results with respect to the defined metrics in \cref{sec:metrics} are detailed in \cref{tab:simu_results} and \cref{tab:real_results} for simulation and real scenarios, respectively.
\par
In the simulation, the monocular configuration performs similarly to the LiDAR, and even better depending on the scenario. For example, Mono-VINS (STCD only) outperforms the LiDAR configuration for the $20$m goal in all simulated environments with an average success rate of $93$\% compared to $67$\% and $63$\% for the LiDAR real-params and sim-tuned, respectively. Upon finer analysis of the recorded trajectories, we found that CSF induces failures, especially for the LiDAR sim-tuned configuration that may explain these results.
This goal is one of the most challenging as it forces the robot to make several turns. The only scenario in which monocular fails is in the hard environment with the $30$m goal, achieving a success rate of $10$\% while the distance ratio is close to one.
This mean that the run failure occurs close to the goal in front of the high grass as shown in \cref{fig:quali_results_simu} for the reasons discussed in \cref{sec:quali}.
\par
Regarding real-world navigation performance, Mono-VINS challenges the LiDAR configuration in all scenarios, achieving the same success rate and distance ratio of 1.0. However, the SPL decreases by 22\% due to the additional distance traveled to achieve the goal as seen in the qualitative results.
Based on those results, the navigation pipeline can be used with only one monocular camera instead of the LiDAR in a real environment while ensuring a robust and accurate navigation with a slight loss of performance or even none at all, depending on the terrain.

\subsection{Output frequencies of the navigation pipeline}\label{sec:exec}

\begin{table}[ht]
    \centering
    \caption{Frequency of each module with the Jetson ORIN (MAXN profile) for the LiDAR-based and Monocular-based configurations}
    \label{tab:exec_time}
    \begin{tabular}{lcc}
        \hline
        \textbf{Module} & \textbf{LiDAR (Hz)} & \textbf{Mono (Hz)} \\
        \hline
        3D perception (CPU/GPU)  & 20 & 10 \\
        Elevation mapping (GPU)      & 16 & 6 \\
        Path planning (CPU)      & 12 & 12 \\
        \hline
    \end{tabular}
\end{table}

\Cref{tab:exec_time} describes the output frequency of data published by each module for the LiDAR and monocular configurations in the Jetson ORIN. The LiDAR-based configuration consistently achieves higher frequencies. The 3D perception module runs at 20 Hz with LiDAR compared to 10 Hz with the monocular setup. This difference is largely due to Depth Anything v2 inference, which introduces significant GPU overhead. The elevation mapping process drops from 16 Hz to 6 Hz in the monocular configuration because the monocular depth estimation leaves less compute available. One of the main improvements to reduce execution time would be to run Depth Anything v2 and Elevation mapping (CSF and the height map) across multiple GPUs.

\section{Ablation Study}

\begin{table}[b]
\centering
\caption{Average SPL performances with different configurations for the depth rescaling}
\label{tab:ablation}
\begin{tabular}{cc|ccc|c}
\hline
\textbf{Edge masking} & \textbf{Smoothing} & \textbf{Easy} & \textbf{Medium} & \textbf{Hard} & \textbf{Avg.} \\ \hline
\multirow{2}{*}{\xmark} & \xmark & 0.50 & 0.48 & 0.19 & 0.39\\
& \cmark  & 0.54 & 0.61 & 0.30 & 0.48\\ \hline
\multirow{2}{*}{\cmark} & \xmark & \textbf{0.63} & 0.74 & 0.18 & 0.52\\ 
& \cmark  & 0.54 & \textbf{0.76} & \textbf{0.45} & \textbf{0.58}\\ \hline
\end{tabular}
\end{table}

\Cref{tab:ablation} analyses the effect of the edge filtering and the smoothing parameter that regularize the scale factor when the visual SLAM system is unstable with the SPL metric. It is the most relevant metric to evaluate the impact of edge filtering and smoothing on overall navigation performance as it measures the navigation efficiency. On average across all scenarios, the activation of both functionalities allows better SPL performances.
It is also noticeable that the configurations using the edge masking consistently outperform those where this feature is not computed. For example, applying smoothing with edge masking yields a 21\% performance improvement over the equivalent configuration without edge masking, demonstrating its importance for achieving better performances. 

\section{CONCLUSION}

This paper introduces a complete navigation stack for ground vehicles operating in off-road environments. The stack processes both point clouds from LiDAR sensors or images through a monocular depth estimation with a foundation model, so no additional training is required.
We have conducted extensive experiments demonstrating robust performance in both complex simulations and unstructured real-world environments.
Our whole stack is made publicly available, providing a training-free pipeline for autonomous off-road navigation, even though it may require parameter tuning, especially to adapt the TEB local planner to a different type of robot.
Our simulation environments are also open-sourced as a challenging benchmark that can be used for rigorous comparison against other navigation pipelines.
However, there remains significant potential for enhancement. 
As shown in the experiments in the most challenging simulation environment, our approach is unable to distinguish high grass from other positive obstacles such as trees or rocks, leading to suboptimal trajectories.
A first improvement step would be to integrate traversability modules capable of classifying high grass as traversable. 
Then, we plan to handle negative obstacles that are a significant challenge due to the absence of information.
Finally, while our navigation stack is currently developed in ROS1, we plan to release a ROS2 version soon to ensure long-term maintainability and facilitate broader adoption within the robotics community.

\balance
\bibliographystyle{IEEEtran}
\bibliography{biblio}

\end{document}